\documentclass[twocolumn, 10pt]{article}

\usepackage[utf8]{inputenc}
\usepackage{geometry}
\usepackage{mathptmx} 
\usepackage{amsmath}  
\usepackage{amssymb}
\usepackage{abstract} 

\usepackage{graphicx}
\usepackage{booktabs} 
\usepackage{float}
\usepackage[font=small,labelfont=bf]{caption} 
\usepackage{tikz} 
\usetikzlibrary{shapes, arrows, positioning}

\usepackage{titlesec}
\usepackage[square,numbers]{natbib}
\usepackage{hyperref} 

\titleformat{\section}{\large\bfseries}{\thesection}{1em}{}
\titleformat{\subsection}{\normalsize\bfseries}{\thesubsection}{1em}{}
\titlespacing*{\section}{0pt}{1.2em}{0.6em}
\titlespacing*{\subsection}{0pt}{1.0em}{0.5em}

\title{\textbf{Augmenting Intelligence: A Hybrid Framework for Scalable and Stable Explanations}}
\author{
    Lawrence Krukrubo, Julius Odede, and Olawande Olusegun \\
    \textit{Department of Computing \& Mathematical Sciences, University of Wolverhampton} \\
    \small\texttt{\{L.Krukrubo@wlv.ac.uk, J.Odede@wlv.ac.uk, O.Olusegun@wlv.ac.uk\}}
}
\date{December, 2025}

\begin{document}

\twocolumn[
  \begin{@twocolumnfalse}
    \maketitle
    \begin{abstract}
        \noindent 
        Current approaches to Explainable AI (XAI) face a ``Scalability-Stability Dilemma.'' 
        Post-hoc methods (e.g., LIME, SHAP) may scale easily but suffer from instability, 
        while supervised explanation frameworks (e.g., TED) offer stability but require 
        prohibitive human effort to label every training instance. 
        This paper proposes a Hybrid LRR-TED framework that addresses this dilemma 
        through a novel ``Asymmetry of Discovery.'' When applied to customer churn prediction, we demonstrate that 
        automated rule learners (GLRM) excel at identifying broad ``Safety Nets'' (retention patterns) but struggle to 
        capture specific ``Risk Traps'' (churn triggers)---a phenomenon we term the \textit{Anna Karenina Principle of Churn}. 
        By initialising the explanation matrix with automated safety rules and augmenting it with a Pareto-optimal 
        set of just \textbf{four} human-defined risk rules, our approach achieves \textbf{94.00\%} predictive accuracy. 
        This configuration outperforms the full 8-rule manual expert baseline while reducing human annotation effort by 50\%, 
        proposing a shift in the paradigm for Human-in-the-Loop AI: 
        moving experts from the role of ``Rule Writers'' to ``Exception Handlers.''
        \vspace{0.8cm} 
    \end{abstract}
  \end{@twocolumnfalse}
]

\section{Introduction}
\label{sec:intro}

\noindent As complex ``black box'' machine learning models become standard in high-stakes domains such as financial 
risk assessment and customer retention, the demand for methods to ensure trust and transparency has grown in 
parallel.\footnote{Reproducible code and experiments are available at: \url{https://github.com/Lawrence-Krukrubo/IBM-Learn-XAI}} 
While post-hoc explanation techniques like LIME \cite{ribeiro2016should} and SHAP \cite{lundberg2017unified} have 
gained popularity, they suffer from a fundamental ``Stability Gap'' \cite{alvarez2018robustness}---small perturbations 
in input data can yield vastly different explanations, undermining user trust.

\noindent To address this, frameworks like Teaching Explanations for Decisions (TED) \cite{hind2019ted} propose an \textit{intrinsic} 
interpretability approach, where models are supervised not just by target labels ($Y$), but by human-provided explanations ($E$). 
While TED inherently supports explanation stability, it introduces a new challenge: the ``Knowledge Bottleneck.'' 
Requiring domain experts to manually annotate every training instance with a specific rationale is unscalable and 
expensive.

\noindent Furthermore, domain experts often exhibit availability bias \cite{duarte2024looking}, causing them to overestimate 
the frequency of rare events. This psychological error manifests as ``Cognitive Overfitting'': the creation of overly 
complex rules for edge cases (``anxiety rules'') that are statistically irrelevant, while missing broader, subtle patterns 
that statistical models excel at detecting. This suggests that a fully manual expert system may suffer from diminishing 
returns, where adding more rules increases complexity without proportionally increasing accuracy.

\paragraph{The Hybrid Hypothesis.}
This paper proposes a Hybrid Linear Rule Regression (LRR) and TED framework---termed \textbf{LRR-TED}---that addresses 
the tension between scalability and stability. We hypothesise that the relationship between Rule Complexity and Model 
Performance follows a Pareto distribution: a small subset of ``high-leverage'' expert rules accounts for the majority 
of the explanatory value, while the remaining ``long tail'' of patterns are more effectively handled by automated rule 
discovery.

\noindent Our contributions are as follows: First, using the customer-churn dataset from the IBM AIX360 package, we 
observe that automated rule learners (GLRM) effectively identify retention ``Safety Nets'' but show limitations in 
capturing specific ``Risk Traps,'' suggesting a structural asymmetry. Second, we introduce a hybrid methodology that 
augments automated discovery with a strategically selected subset of domain expert constraints.

\noindent Finally, we present empirical results showing that a Hybrid model with just \textbf{four rules} achieves \textbf{94.00\% accuracy}, 
not only outperforming the fully automated baseline but also exceeding the full 8-rule manual expert system. This 
suggests that restricting human input to high-level strategic constraints can yield better results than exhaustive 
manual rule-writing.
\section{Related Work}
\label{sec:related_work}

\paragraph{The Challenge of Post-Hoc Explanations.}
Widely used post-hoc methods like LIME \cite{ribeiro2016should} and SHAP \cite{lundberg2017unified} rely on local surrogates to 
approximate complex model behaviours. While this approach scales effectively to black-box models, it presents a significant 
challenge: it explains the \textit{prediction} rather than the \textit{model}, often resulting in varying fidelity to the true decision 
boundary. As demonstrated by Alvarez-Melis and Jaakkola \cite{alvarez2018robustness}, these local approximations can be susceptible to 
instability---even infinitesimal input perturbations can trigger vastly different explanations. This resulting ``Stability Gap'' 
makes them less suitable for high-stakes environments, where consistent reasoning is a critical requirement for trust.

\paragraph{Intrinsic Interpretability.}
In response to the stability issues in post-hoc methods, a strategic shift towards \textit{intrinsic interpretability} 
has gained traction. This paradigm eschews post-hoc analysis in favour of models whose internal structure---be it a set of linear 
coefficients, decision rules, or tree splits---is intelligible and directly constitutes the explanation. A prominent 
example of this approach is the Teaching Explanations for Decisions (TED) framework \cite{hind2019ted}. TED learns not only from 
labels but also from human-provided rationales, designed to improve stability and fidelity. However, its reliance on manual 
annotation creates a \textbf{``Knowledge Bottleneck,''} making it challenging to scale for industrial applications where 
annotating every training instance is cost-prohibitive.

\paragraph{Automated Rule Discovery.}
As an alternative, pure rule-based systems like Generalised Linear Rule Models (GLRM) \cite{wei2019generalized} or Boolean decision 
rules \cite{dash2018boolean} offer automated transparency. These models generate explicit IF-THEN rules, removing the 
manual annotation bottleneck. However, this simplicity often comes at a cost: rule-based models may be limited in their ability to capture complex, 
non-linear relationships, potentially leading to lower predictive accuracy compared to deep learning benchmarks. Furthermore, standard 
algorithms often optimise for global coverage, which can mask subtle, minority-class patterns \cite{weiss2004mining} or 
limit the distinction between retention and churn factors---a phenomenon exacerbated by class imbalance \cite{he2009learning}.

\paragraph{Bridging the Gap.}
The existing landscape presents a clear dichotomy: intrinsic methods (TED) offer stability but face scalability hurdles, while automated 
methods (GLRM) offer scalability but often sacrifice accuracy. Our work aims to bridge this gap by adopting a \textit{Human-in-the-Loop} 
strategy. We use automated rule discovery to generate the initial ``Safety'' patterns (automating the supervision), while integrating 
expert domain knowledge to handle specific ``Risk'' exceptions. This hybrid framework seeks to balance the efficiency of data-driven 
models with the reliability of human expertise.
\section{Methodology}
\label{sec:methodology}

\noindent We propose a hybrid framework that integrates the automated structural discovery of rule-based models with the 
high-fidelity supervision of the Teaching Explanations for Decisions (TED) framework. Our pipeline operates in three distinct 
phases: (1) Automated Discovery of retention structures (``Safety Nets''); (2) Domain Augmentation to define specific churn 
triggers (``Risk Traps''); and (3) the training of a hybrid TED-SVC classifier.

\paragraph{Experimental Setup.}
To validate our framework, we utilised the Customer Churn Dataset from the IBM AIX360 interpretability demo \cite{arya2019one}. 
This dataset represents a classic high-stakes classification problem where the goal is to predict customer attrition based on 
demographic and behavioural attributes. Before modelling, we performed standard preprocessing steps, including the removal of 
unique identifiers. As a prerequisite for rule-based discovery, we transformed the continuous and categorical features into a 
high-dimensional binary feature space using the \textit{FeatureBinarizer} module from the AIX360 toolkit. This process, based on 
the column generation technique by Dash et al. \cite{dash2018boolean}, creates atomic propositions (e.g., \textit{Age $<$ 30}, \textit{Tenure $\ge$ 5}) and 
their negations. This binarisation is a prerequisite for the linear rule regression phase, allowing the model to assign weights to 
explicit logical conditions rather than raw numerical values.

\begin{figure*}[t!]
    \centering
    \includegraphics[width=0.825\textwidth]{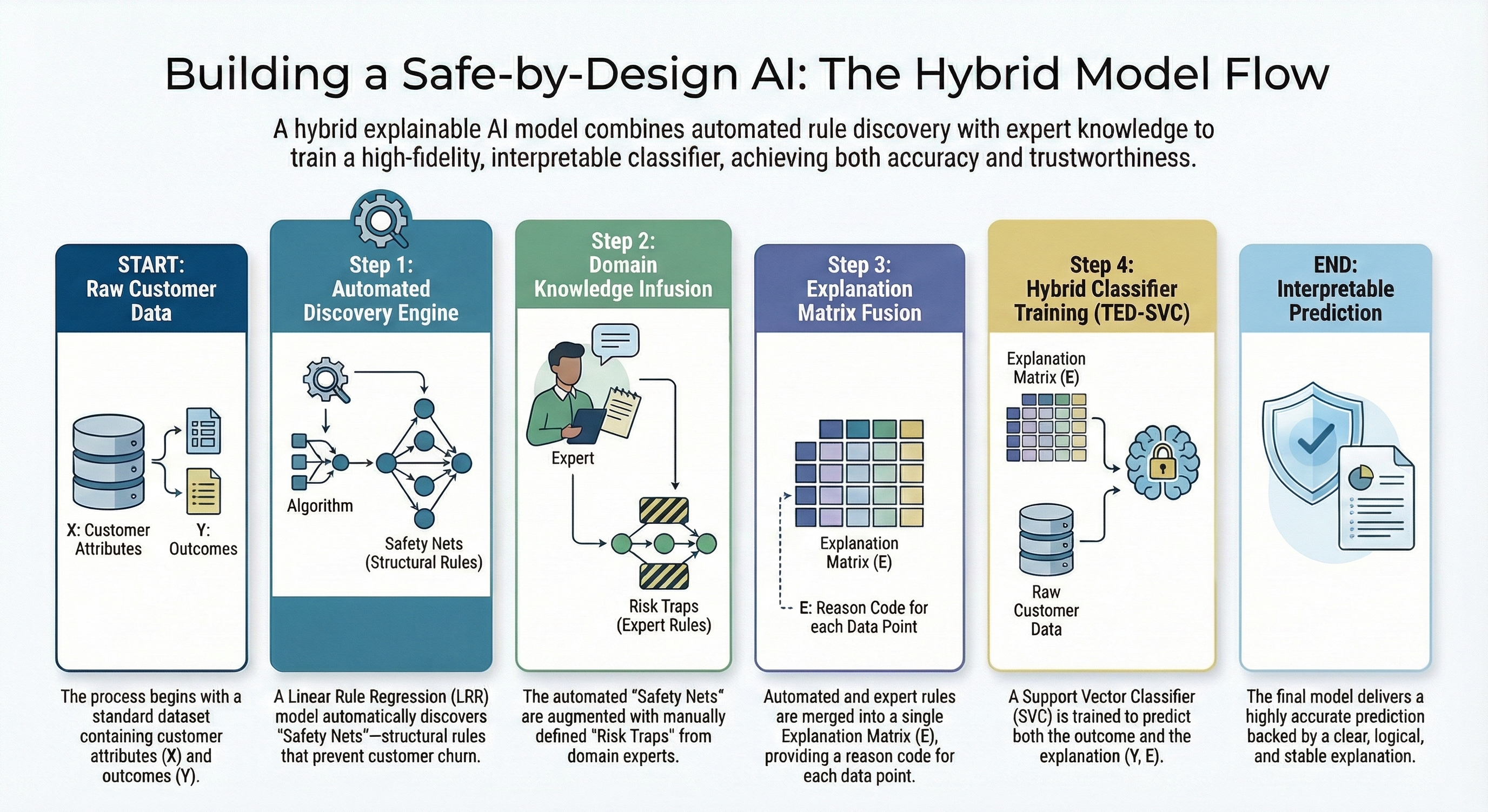} 
    \caption{\textbf{The Hybrid LRR-TED Pipeline.} Automated rules (Safety) and Expert rules (Risk) are fused into a single 
    Explanation Matrix ($E$) to initialise the supervised classifier. The process moves from broad statistical discovery to 
    granular expert refinement.}
    \label{fig:pipeline}
\end{figure*}

\paragraph{Phase 1: Automated Discovery (LRR).}
To address the ``Knowledge Bottleneck'' of manual rule creation, we employ an unsupervised discovery phase to extract latent 
logic from the training data. We utilise the Generalised Linear Rule Model (GLRM) as proposed by Wei et al. \cite{wei2019generalized} 
and apply Linear Rule Regression (LRR) to the binarised feature set $\mathbf{B}$. LRR optimises a sparse linear model where the 
coefficient $w_j$ represents the contribution of rule $r_j$ to the prediction target $Y$ (Churn). The optimisation objective is
\begin{equation}
    \min_{\mathbf{w}} \frac{1}{2} \|\mathbf{Y} - \mathbf{B}\mathbf{w}\|^2_2 + \lambda_1 \|\mathbf{w}\|_1 + \lambda_2 \sum_j C(r_j)
    \label{eq:lrr_objective}
\end{equation}
where $C(r_j)$ represents the complexity cost of rule $j$.

\noindent \textbf{Proposed Approach: Safety Tiers.} We formalise the discovery of ``Safety Nets'' as the set of rules $S = \{r_j : w_j < 0\}$, where $w_j$ is 
the learned coefficient from the GLRM. Rather than mapping each rule to a unique explanation, we aggregated them into three 
hierarchical classes (Codes 1--3) based on the magnitude of their coefficients $|w_j|$. Code 1 encapsulates the strongest 
retention drivers, while Codes 2 and 3 capture moderate and secondary safety factors, respectively. This aggregation prevents 
explanation sparsity and provides a tiered confidence level for why a customer is ``Safe.''

\paragraph{Phase 2: Domain Augmentation (Risk Filter).}
While LRR effectively identifies broad retention patterns, it can lack the semantic precision required to identify specific, 
actionable ``Risk Traps'' (e.g., \textit{Payment Delay $>$ 20 days}). To address this, we augment the automated Safety Nets with 
confirmatory Domain Knowledge. We construct a composite Explanation Matrix $\mathbf{E}$ by merging the two logic streams:
\begin{itemize}
    \item \textbf{Safety Nets (Automated):} Customers satisfying LRR retention rules are assigned Codes 1--3.
    \item \textbf{Risk Traps (Manual):} Customers satisfying domain-defined risk predicates are assigned Codes 4--11.
    \item \textbf{The Default State (Code 12):} We introduce a ``Drift'' class for observations that trigger neither a Safety Net 
    nor a Risk Trap. This captures the intersection of ``No Safety'' and ``No Specific Risk,'' often representing a pre-churn 
    state invisible to standard classifiers.
\end{itemize}

\paragraph{Strategic Rule Selection (Pareto Filter).}
To bridge the gap between the manual expert system (8 rules) and the automated baseline, we employed a \textit{Pareto-based selection strategy} 
to identify the minimal set of domain rules required to maximise performance. The full set of 8 manual rules was analysed along 
two dimensions:
\begin{enumerate}
    \item \textbf{Coverage (Volume):} The percentage of the churn population captured by the rule. Rules capturing $<1\%$ of the population were deemed candidates for automated (LRR) handling.
    \item \textbf{Orthogonality (Uniqueness):} To minimise redundancy, we validated the independence of the selected rules using the Jaccard Similarity metric:
    \begin{equation}
        J(A,B) = \frac{|A \cap B|}{|A \cup B|}
        \label{eq:jaccard}
    \end{equation}
    Our analysis indicated a high degree of orthogonality among the ``Golden Quartet,'' with an average pairwise similarity of 
    just \textbf{0.09} ($9\%$) and a maximum overlap of $\textbf{0.26}$.
\end{enumerate}
Based on this analysis, we down-selected the expert logic to a ``Golden Quartet'' of four rules focusing on distinct behavioural 
quadrants: \textbf{Financial Risk}, \textbf{Structural Risk}, \textbf{Interaction Risk}, and \textbf{Engagement Risk}.

\paragraph{Phase 3: The Hybrid Classifier (TED-SVC).}
The final phase utilises the Teaching Explanations for Decisions (TED) framework \cite{hind2019ted} to train a high-accuracy 
Support Vector Classifier (SVC). Unlike standard supervision which maps $\mathbf{X} \rightarrow \mathbf{Y}$, TED maps $\mathbf{X} \rightarrow (\mathbf{Y}, \mathbf{E})$. 
We construct the Cartesian product of the label space and the explanation space, creating a set of unique class identifiers $Y \times E$. The SVC is trained to minimise a joint loss function that enforces fidelity to both the label and the explanation:
\begin{equation}
    \mathcal{L} = \sum_{i=1}^{N} (L(y_i, \hat{y}_i) + \mu L(e_i, \hat{e}_i))
    \label{eq:ted_loss}
\end{equation}
By initialising the explanation vector $e_i$ with our hybrid rules (Safety Nets + Risk Traps), we force the SVC to learn decision 
boundaries that align with valid domain logic. This results in a model that is ``Safe-by-Design,'' achieving high predictive 
accuracy while ensuring that every prediction is backed by either a discovered structural rule or a verified domain constraint.
\section{Results}
\label{sec:results}

\noindent We evaluated the proposed Hybrid framework against two control conditions: a fully manual baseline (representing the ideal but unscalable ``Gold Standard'') 
and a fully automated baseline (representing the state-of-the-art in rule discovery). Our experiments aimed to quantify 
the trade-off between human effort (rule complexity) and model fidelity (Y+E Accuracy).

\begin{table*}[t!] 
\centering
\caption{The Efficiency Frontier: Evolution of Model Performance. The Hybrid (4-Rule) configuration exceeds the manual expert 
benchmark while halving the complexity.}
\label{tab:efficiency_frontier}
\setlength{\tabcolsep}{10pt} 
\begin{tabular}{@{}lccl@{}}
\toprule
\textbf{Model Configuration} & \textbf{Rule Count} & \textbf{Y+E Accuracy} & \textbf{Narrative Role} \\ \midrule
LRR (Full Automation)       & 0                   & 75.15\%               & Baseline (No Human Input) \\
Hybrid (Behavioural Trio)   & 3                   & 90.05\%               & Efficiency Champion \\
Manual TED (Benchmark)      & 8                   & 92.90\%               & Human Gold Standard \\
\textbf{Hybrid (Golden 4)}  & \textbf{4}          & \textbf{94.00\%}      & \textbf{Best Configuration} \\ \bottomrule
\end{tabular}
\end{table*}

\noindent The performance metrics for the four model configurations are summarised in Table \ref{tab:efficiency_frontier}.

\paragraph{The Automation Gap.}
The fully automated Generalised Linear Rule Model (GLRM) established a baseline combined accuracy of 75.15\%. While the model 
successfully identified churn predictors, it showed limited alignment with specific domain explanations, exhibiting the ``Code 12'' 
phenomenon where risky customers were unidentified by broad structural rules. By injecting just three behavioural rules (Payment, Spend, Support) 
into the explanation matrix, the Hybrid model achieved a significant 14.9\% increase in accuracy to 90.05\%. This suggests that domain 
expertise is likely most valuable when creating ``guardrails'' for the AI, rather than attempting to define every rule manually.

\begin{figure}[t!] 
\centering
\includegraphics[width=\columnwidth]{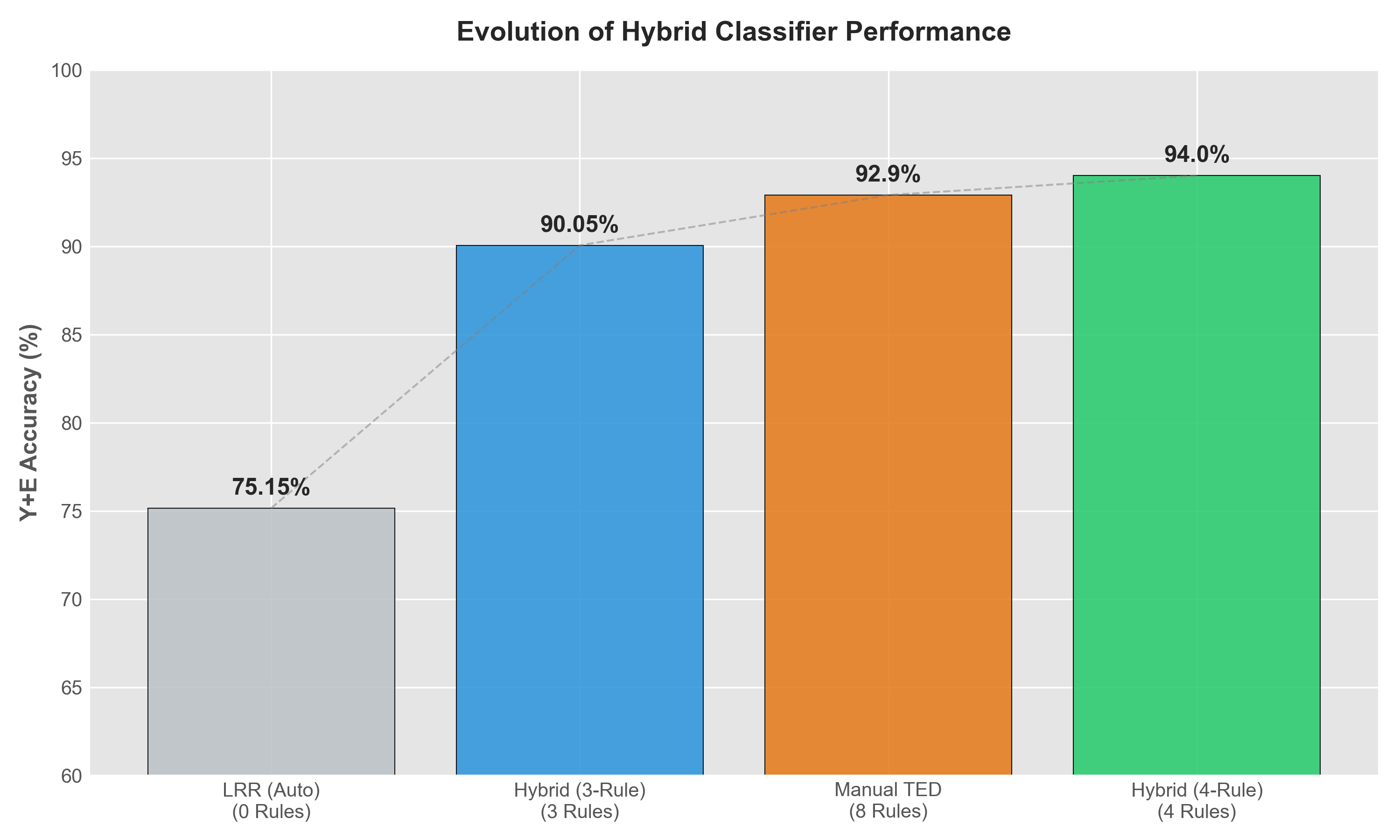}
\caption{The Efficiency Frontier: A comparison of rule complexity versus combined accuracy. The Hybrid approach (Blue/Green) 
demonstrates a steep ascent in accuracy with minimal rules, eventually crossing the Manual Benchmark (Orange).}
\label{fig:efficiency_frontier}
\end{figure}

\paragraph{The Efficiency Frontier.}
Remarkably, our 3-rule Hybrid model achieved 97\% of the performance of the full 8-rule Manual system (90.05\% vs 92.90\%). This 
suggests that a substantial portion of the ``expert value'' in churn prediction is concentrated in a few key behaviours---specifically 
payment delinquency and support interactions---and that the remaining manual rules may yield diminishing returns. This configuration 
represents a robust choice for resource-constrained environments, delivering near-expert performance with minimal annotation 
effort.

\paragraph{Exceeding the Benchmark.}
By adding a fourth structural rule (`Monthly Contract') to explicitly handle structural churn risks, the Hybrid model achieved \textbf{94.00\% accuracy}. 
This setup outperforms the full 8-rule Manual expert system by 1.1\%, while commendably using 50\% fewer rules. This 
suggests that a Hybrid AI system is not merely a compromise between automation and accuracy, but can offer advantages over a 
fully manual system by combining the consistency of AI with the semantic precision of human experts.

\begin{table}[h!] 
\centering
\caption{Detailed Classification Report. The high Precision (0.99) for Churn minimises false positives, while the Recall (0.93) 
indicates the 4 rules successfully capture the ``Risk Traps.''}
\label{tab:classification_report}
\setlength{\tabcolsep}{4pt} 
\begin{tabular}{@{}lcccc@{}}
\toprule
\textbf{Class} & \textbf{Precision} & \textbf{Recall} & \textbf{F1-Score} & \textbf{Support} \\ \midrule
0 (Stay)       & 0.92               & 0.98            & 0.95              & 860              \\
1 (Churn)      & 0.99               & 0.93            & 0.96              & 1140             \\ \midrule
\textbf{Wt. Avg} & \textbf{0.96}      & \textbf{0.95}   & \textbf{0.95}     & \textbf{2000}    \\ \bottomrule
\end{tabular}
\end{table}

\paragraph{Component Analysis (Precision \& Recall).}
Beyond the combined $Y+E$ accuracy, we analysed the pure predictive performance of the Hybrid model (Table \ref{tab:classification_report}) 
to assess its business utility. The model achieved a near-perfect \textbf{Precision of 0.99} for the Churn class, indicating that 
false positives are significantly minimised---a critical factor for optimising retention budgets. Simultaneously, the \textbf{Recall of 0.93} 
indicates that our ``Golden Quartet'' of rules successfully captured the vast majority of at-risk customers, suggesting that the 4-rule set 
provides sufficient coverage of the risk landscape without the need for the redundant complexity of the full 8-rule manual system.
\section{Discussion}
\label{sec:discussion}

\noindent While our results suggest that a Hybrid TED framework can outperform manual expert systems with significantly 
reduced effort, deeper analysis reveals important structural dynamics and boundary conditions regarding the generalisability of 
this approach.

\paragraph{The Anna Karenina Principle.}
A counter-intuitive finding emerged from our structural analysis. Although Churn was the majority class in our training sample, 
the automated LRR algorithm primarily discovered ``Safety Nets'' (retention rules) while showing limited capacity to identify ``Risk Traps'' (churn rules). 
This suggests a deeper structural asymmetry in customer behaviour, which we term the \textit{Anna Karenina Principle of Churn}: 
satisfied customers appear to exhibit broad, homogeneous behaviours (dense clusters), whereas churning customers leave for highly specific, 
heterogeneous reasons (sparse disjoints). Because the automated learner optimises for simplicity, it naturally gravitates towards the 
broad patterns of retention. This supports the premise that human expertise is essential not to label the ``normal'' data, but 
to identify the specific ``long tail'' of risks that automated tools may overlook.

\paragraph{Limitation: Class Imbalance.}
A notable observation of this study is the class distribution of the experimental dataset. Our training data contained a 
relatively balanced churn rate ($\approx 56\%$). In contrast, many real-world production environments exhibit extreme 
class imbalance, where churn events are rare anomalies ($<5\%$). It remains to be investigated whether the LRR's ability to discover ``Safety Nets'' 
would degrade in a highly imbalanced environment where the ``Stay'' class is overwhelmingly dominant. We hypothesise that in such 
scenarios, the automated model might default to a trivial ``always stay'' rule (The Null Paradox), potentially necessitating the human 
expert to define $\approx 100\%$ of the churn logic manually.

\paragraph{Limitation: Linearity vs. Complexity.}
Our reliance on Linear Rule Regression (LRR) assumes that the underlying decision boundaries can be approximated by rectangular hyper-boxes (Boolean rules). 
While effective for tabular business data (e.g., Telecom, Finance), this approach may not generalise to domains with highly non-linear 
or perceptual features, such as image recognition or natural language sentiment analysis, where feature binarisation results in significant 
information loss.

\paragraph{Future Work.}
Future research should focus on stress-testing the ``Efficiency Frontier'' across diverse data distributions. Specifically, we 
propose evaluating the Hybrid LRR-TED framework on datasets with varying imbalance ratios (from 1:1 to 1:100) to determine the ``break-even point'' 
where automated rule discovery diminishes in utility. Additionally, integrating causal inference techniques could help distinguish 
between correlational safety nets (e.g., ``High Usage'') and causal retention drivers, further enhancing the stability of the generated explanations.
\section{Conclusion}
\label{sec:conclusion}

\noindent This paper presented a novel Hybrid LRR-TED framework that addresses the scalability-stability dilemma in Explainable AI. 
By utilising automated rule discovery to initialise explanations and augmenting them with a Pareto-based set of human domain 
constraints, we achieved a combined predictive and explanatory accuracy of \textbf{94.00\%}, exceeding the full manual expert 
benchmark while reducing annotation effort by 50\%.

\noindent Our findings support a proposed shift in the role of the domain expert: from a ``Rule Writer'' who must exhaustively define the 
world, to an ``Exception Handler'' who creates strategic guardrails for an automated learner. As AI systems become increasingly 
ubiquitous in high-stakes decision-making, such hybrid frameworks offer a pragmatic path toward models that are scalable, accurate, 
and inherently interpretable.

\bibliographystyle{plainnat}
\bibliography{references}

\begin{thebibliography}{10}
\providecommand{\natexlab}[1]{#1}
\providecommand{\url}[1]{\texttt{#1}}
\expandafter\ifx\csname urlstyle\endcsname\relax
  \providecommand{\doi}[1]{doi: #1}\else
  \providecommand{\doi}{doi: \begingroup \urlstyle{rm}\Url}\fi

\bibitem[Alvarez-Melis and Jaakkola(2018)]{alvarez2018robustness}
David Alvarez-Melis and Tommi~S Jaakkola.
\newblock On the robustness of interpretability methods.
\newblock In \emph{Proceedings of the 2018 Workshop on Human Interpretability
  in Machine Learning}, 2018.

\bibitem[Arya et~al.(2019)Arya, Bellamy, Chen, Dhurandhar, Hind, Hoffman,
  Houde, et~al.]{arya2019one}
Vijay Arya, Rachel K~E Bellamy, Pin-Yu Chen, Amit Dhurandhar, Michael Hind,
  Samuel~C Hoffman, Stephanie Houde, et~al.
\newblock One explanation does not fit all: A toolkit and taxonomy of
  explainable {AI} policies.
\newblock \emph{arXiv preprint arXiv:1909.03012}, 2019.

\bibitem[Dash et~al.(2018)Dash, G{\"u}nl{\"u}k, and Wei]{dash2018boolean}
Sanjeeb Dash, Oktay G{\"u}nl{\"u}k, and Dennis Wei.
\newblock Boolean decision rules via column generation.
\newblock In \emph{Advances in Neural Information Processing Systems},
  volume~31, 2018.

\bibitem[Duarte and Campos(2024)]{duarte2024looking}
Regina de~Brito Duarte and Joana Campos.
\newblock Looking for cognitive bias in {AI}-assisted decision-making.
\newblock In \emph{Proceedings of the HHAI 2024 Workshops}, volume 3825, pages
  16--22. CEUR-WS.org, 2024.

\bibitem[He and Garcia(2009)]{he2009learning}
Haibo He and Edward~A Garcia.
\newblock Learning from imbalanced data.
\newblock \emph{IEEE Transactions on Knowledge and Data Engineering},
  21\penalty0 (9):\penalty0 1263--1284, 2009.

\bibitem[Hind et~al.(2019)Hind, Wei, Campbell, Codella, Dhurandhar,
  Mojsilovi{\'c}, Ramamurthy, and Varshney]{hind2019ted}
Michael Hind, Dennis Wei, Murray Campbell, Noel C.~F. Codella, Amit Dhurandhar,
  Aleksandra Mojsilovi{\'c}, Karthikeyan~Natesan Ramamurthy, and Kush~R.
  Varshney.
\newblock {TED}: Teaching {AI} to explain its decisions.
\newblock In \emph{Proceedings of the 2019 {AAAI}/{ACM} Conference on {AI},
  Ethics, and Society}, pages 123--129, 2019.

\bibitem[Lundberg and Lee(2017)]{lundberg2017unified}
Scott~M Lundberg and Su-In Lee.
\newblock A unified approach to interpreting model predictions.
\newblock In \emph{Advances in Neural Information Processing Systems},
  volume~30, 2017.

\bibitem[Ribeiro et~al.(2016)Ribeiro, Singh, and Guestrin]{ribeiro2016should}
Marco~Tulio Ribeiro, Sameer Singh, and Carlos Guestrin.
\newblock "why should i trust you?": Explaining the predictions of any
  classifier.
\newblock In \emph{Proceedings of the 22nd ACM SIGKDD International Conference
  on Knowledge Discovery and Data Mining}, pages 1135--1144, 2016.

\bibitem[Wei et~al.(2019)Wei, Dash, Gao, G{\"u}nl{\"u}k, Watson, and
  Wei]{wei2019generalized}
Dennis Wei, Sanjeeb Dash, Tian Gao, Oktay G{\"u}nl{\"u}k, Barnabas Watson, and
  Dennis Wei.
\newblock Generalized {Linear} {Rule} {Models}.
\newblock In \emph{International Conference on Machine Learning}, pages
  6687--6696. PMLR, 2019.

\bibitem[Weiss(2004)]{weiss2004mining}
Gary~M Weiss.
\newblock Mining with rarity: A unifying framework.
\newblock \emph{ACM SIGKDD Explorations Newsletter}, 6\penalty0 (1):\penalty0
  7--19, 2004.

\end{thebibliography}

\end{document}